\documentclass{ifacconf}

\usepackage{natbib}        
\usepackage{color,soul}
\usepackage{amsmath,amssymb,amsfonts}
\usepackage{textcomp}
\usepackage{subfigure}
\usepackage{bm}
\usepackage{verbatim}
\usepackage{graphicx}
\usepackage[thinlines,thiklines]{easybmat}
\usepackage{latexsym}
\usepackage{algorithm}
\usepackage{algpseudocode}

\def\s{\mathop{\rm s}\nolimits}
\def\c{\mathop{\rm c}\nolimits}
\newcommand{\bs}[1]{\ensuremath{{\boldsymbol{#1}}}}

\def\atan2{\mathrm{atan2}}

\def\mathbi#1{\textbf{\em #1}}

\begin{document}
\begin{frontmatter}

\title{Energy Efficient Foot-Shape Design for Bipedal Walkers on Granular Terrain}

\thanks[footnoteinfo]{Address all correspondence to J. Yi.}

\author[First]{Xunjie Chen}
\author[First]{Jingang Yi}
\author[Second]{Hao Wang}

\address[First]{Department of Mechanical and Aerospace Engineering \\ Rutgers University, Piscataway, NJ 08854 USA \\ (e-mail: \{xc337, jgyi\}@rutgers.edu)}
\address[Second]{Department of Civil and Environment Engineering \\ Rutgers University, Piscataway, NJ 08854 USA \\ (e-mail: hw261@soe.rutgers.edu)}

\begin{abstract}
It is important to understand how bipedal walkers balance and walk effectively on granular materials, such as sand and loose dirt, etc. This paper first presents a computational approach to obtain the motion and energy analysis of bipedal walkers on granular terrains and then discusses an optimization method for the robot foot-shape contour design for energy efficiently walking. We first present the foot-terrain interaction characteristics of the intrusion process using the resistive force theory that provides comprehensive force laws. Using human gait profiles, we compute and compare the ground reaction forces and the external work for walking gaits with various foot shapes on granular terrains. A multi-objective optimization problem is finally formulated for the foot contour design considering energy saving and walking efficiency. It is interesting to find out a non-convex foot shape gives the best performance in term of energy and locomotion efficiency on hard granular terrains. The presented work provides an enabling tool to further understand and design efficient and effective bipedal walkers on granular terrains.
\end{abstract}

\begin{keyword}
Bipedal walkers, granular terrains, foot-ground interactions, gait design
\end{keyword}

\end{frontmatter}

\section{Introduction}

Bipedal robots appear advantageous for locomotion in complex uneven and deformable, granular terrain~\citep{xiong2017stability}. However, the physical modeling of the granular intrusion interactions complicates the walking dynamics because unlike the rigid flat ground, there is lack of the comprehensive force laws for foot-granular terrain contact. Although there are applications of wheel-based robot locomotion to investigate terradynamic performance on granular materials such as sand, few studies focus on humanoid bipedal foot-granular terrain interaction and corresponding influence on walking dynamics and performance.

Bipedal walking gaits has been investigated from the energy expenditure view. Inverted pendulum model with rocker feet has proved that rolling feet with a certain curvature leads to an advantage for reducing step-to-step transition work and saving energy~\citep{adamczyk2006advantages,TrkovJCND2019}. Complex curved foot shape strongly affects performance and efficiency in bipedal walking. Properties (e.g., radius) of rocker feet are investigated to show the effect on the energetic efficiency of a gait and the hip kinematics~\citep{martin2011experimental,martin2012effects}.
However, all of the above-mentioned work focus on the robot walking on the rigid flat ground and locomotion on the granular materials are much complicated.

Although conventional terramechanics reveals good wheel sinkage estimation of off-road vehicles on soil, legged robots face a more complicated scenario. Large-scale numerical computation algorithms such as discrete element method and material point method can predict forces accurately for diverse movements~\citep{dunatunga2017continuum,wang2021continuum}. However, it is impractical to use these methods for real-time estimation and robot control applications. An alternative computation method, namely resistive force theory (RFT), was used to capture the interaction of arbitrary shapes with dry granular materials for 2D and 3D trajectories~\citep{li2013terradynamics,TreersRAL2021}. RFT has already been successfully applied to calculate reaction forces for wheels rotating on sand~\citep{agarwal2019modeling,agarwal2021surprising,Huang2022RAL} and generalized for similar dry materials by introducing a scale factor~\citep{zhang2014effectiveness}.
One static intrusion reaction force model was used to rationalize a static stability region criterion for the inverted pendulum model on granular materials~\citep{xiong2017stability}.
Nevertheless, few works focus on complex foot shape impact on granular materials following resistive force theory principles.

In this paper, we follow the general rules of RFT to quantify the reaction forces during the interaction of robot feet with dry granular terrains. We present an optimization approach for robot foot shapes by considering the compensation energy and walking efficiency. The intrusion orientation and motion direction of any arbitrary point on the foot contour are calculated. A computational method is proposed to calculate the ground resistive forces on walking dynamics by RFT. Various commonly used foot shapes are used as examples to analyze and compare the robot walking performance. Finally, we optimize the locations of finite waypoints to identify and estimate the foot shape that generates energy-saving and efficient walking performance. Different types of granular materials are also tested for the optimized foot shape. It is found that the optimized foot shape is non-convex for energy-efficient walking gaits on hard granular terrains. The main contribution of this work lies in that the integrated bipedal walking gait dynamics with the resistive force theory for studying foot-granular terrain interactions. The proposed integration foot-ground modeling framework is not restricted for only minimizing energy efficiency for bipedal walking on granular materials, and it indeed lays down the foundation for future work such as balance and gait control for bipedal robotic walkers on deformable, granular terrains.

The rest of this paper is organized as follows. Section~\ref{sec:RFTsimullation} presents a computational method that integrates the RFT for the robot walking on the granular terrain. Section~\ref{sec:walkingAnalysis} discusses the walking energy efficiency with various foot shapes on granular terrain. We present the foot shape optimization in Section~\ref{sec:optimalDesign}. Finally, results and conclusions are presented in Sections~\ref{sec:Results} and~\ref{sec:conclusion}, respectively.

\section{Bipedal Dynamics on Granular Terrains}
\label{sec:RFTsimullation}
\subsection{Robot Foot Intrusion Forces through RFT}
\begin{figure}[t!]
    \centering
    \includegraphics[width=3.0in]{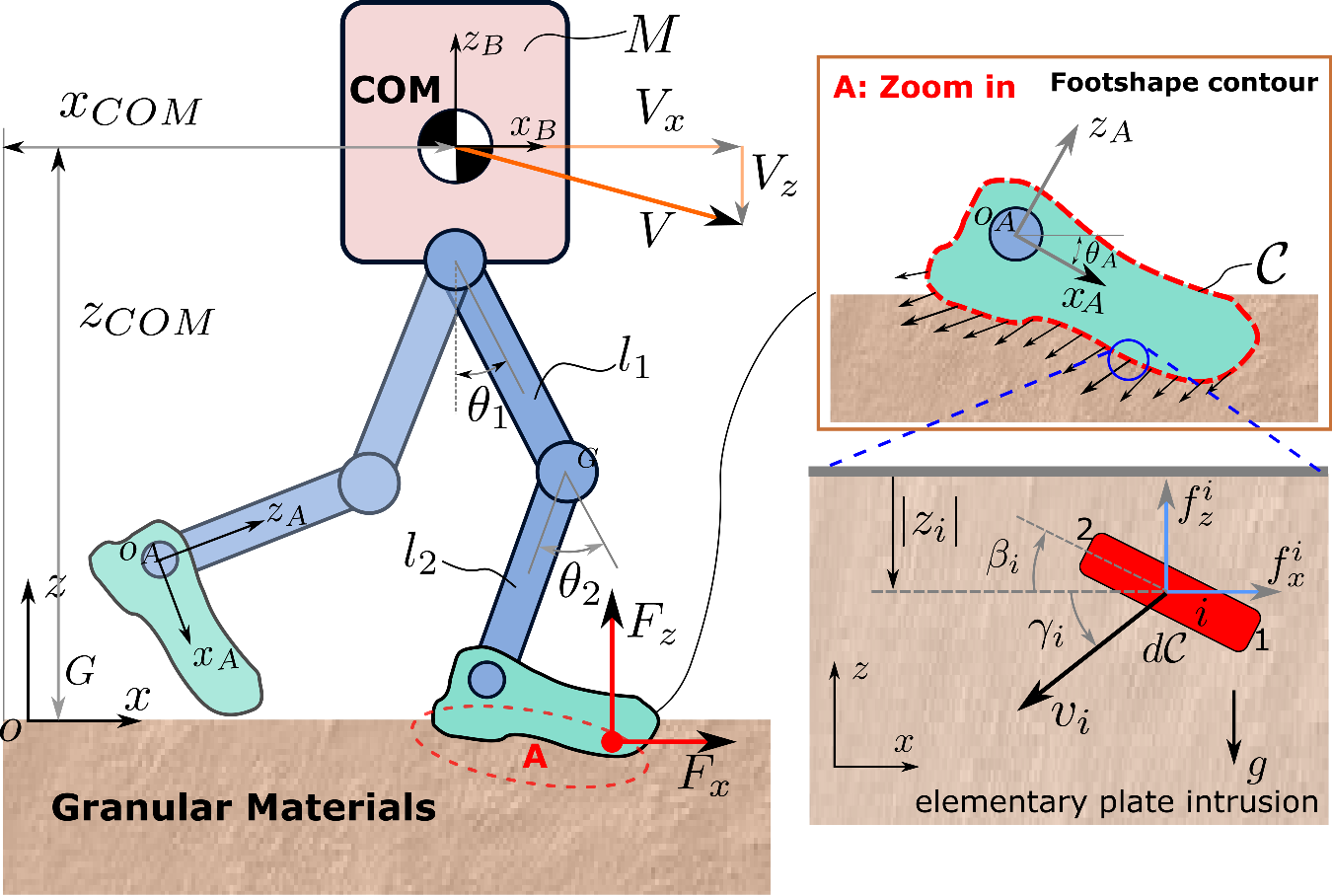}
 	\caption{The schematic of a humanoid robot walking and the arbitrary-shaped foot intrusion with granular materials. RFT is illustrated with discrete elementary plates.}
    \label{fig:RobotWalkingSchematic}
\end{figure}

Fig.~\ref{fig:RobotWalkingSchematic} illustrates the systems configuration of the bipedal walker on granular terrains. A double-link and one-lumped-mass model is used to present the motion of the robotic walker which is in the sagittal plane. Besides the inertial frame ($G:\,xoz$), a local body-fixed frame ($B:x_Bo_Bz_B$) at the center of mass (COM) and a local ankle frame ($A:x_Ao_Az_A$) are established to describe relative positions and velocities. The double-link legs are symmetric and the lengths of the thing and shank links are denoted as $l_1$ and $l_2$, respectively. The hip (with the vertical line) and knee angles are denoted as $\theta_1$ and $\theta_2$, respectively.

The robot foot intrusion problem is also illustrated in Fig.~\ref{fig:RobotWalkingSchematic}. To simplify the problem, the foot is considered as rigid and the RFT rules are applied to intrusion of vertical plates allowing asymmetric stress distributions. The resistive forces are also considered to be uniformly distributed in the foot width direction so that the intrusion forces and motion of interest are only considered in the $x$-$z$ plane. It is further assumed that the intrusion forces are not sensitive to foot gait velocity under a slow interaction motion.

We consider the robotic foot with an arbitrary out-shape contour $\mathcal{C}$. The entire contour $\mathcal{C}$ is partitioned into an $N$ elementary plates with the same length $d\mathcal{C}$. The width of each plate is negligible. For the $i$th plate, according to RFT, the drag and lift forces of intrusion parts, denoted respectively as $f_{x}^{i}$ and $f_{z}^{i}$, are approximately proportional to the intrusion depth, $|z|$ (free level of granular materials is $z=0$) as
\begin{equation}
\label{eqn:RFT}
f_{j}^{i} = \alpha_{j}(\beta_i , \gamma_i)|z_i| H(-z_i), \;j=x,z,
\end{equation}
where $i=1,\ldots,N$. $\alpha_{x}$ and $\alpha_{z}$ reflect local stresses in the horizonal and vertical direction per unit depth, respectively. $H(x)$ is Heaviside function. As shown in Fig.~\ref{fig:RobotWalkingSchematic}, for the $i$th plate, the orientation angle, $\beta_i$ is defined clockwise with respect to $x$-axis while the plate moves towards a certain direction described by a velocity vector angle $\gamma_i$. The total horizontal and vertical forces applied on feet are
\begin{equation}
\label{eqn:RFT_totalforce}
F_{j} = \sum_{i=1}^N \alpha_{j}(\beta_i, \gamma_i)|z_i| H(-z_i), \; j=x,z.
\end{equation}
The local horizontal and vertical stresses $\alpha_{x}$ and $\alpha_{z}$ are obtained by small plate intrusion experiments and gathered as two 2-D force lookup maps with respect to intrusion orientations ($\beta_i$) and motion directions ($\gamma_i$). Given a foot contour and a walking gait, the plate orientation and motion direction are calculated as
\begin{equation}
\label{eqn:beta_gamma}
\beta_i= \tan^{-1}{\left(\frac{z_{i2}-z_{i1}}{x_{i2}-x_{i1}}\right)}, \gamma_i=\tan^{-1}{\left(\frac{v_{iz}}{v_{ix}}\right)},
\end{equation}
where $(x_{i1},z_{i1})$ and $(x_{i2},z_{i2})$ are the two vertices coordinates of the $i$th plate and $\mathbi{v}_i=[v_{ix} \;v_{iz}]^T$ is the velocity at the center of each plate in the inertial frame ($xoz$). For humanoid walking gaits, the stance foot rolls on the terrain and swing foot touches down with a low relative velocity with respect to the COM. Hence, the intrusion velocity is in a small value region where the local stresses are not sensitive to the velocity \citep{ding2011drag,li2013terradynamics}. We do not consider cases with high intrusion velocity such as in running gaits in this paper.

\subsection{Kinematics and Kinetics of Bipedal Walker}

\subsubsection*{Kinematics}
For an arbitrary point $p_i$ of the foot, the position and corresponding velocity in the global frame are calculated as
\begin{equation}\label{eqn:anypointPosition}
^{G}\mathbi{r}_{p_i} =~^{G}\mathbf{R}^{A}~^{A}\mathbi{r}_{p_i}+~^{G}\mathbi{r}_A,~^{G}\mathbi{v}_{p_i}= \dot{^{G}\mathbf{R}^{A}}~^{A}\mathbi{r}_{p_i}+~^{G}\mathbi{v}_A,
\end{equation}
where $^{A}\mathbi{r}_{p_i}$ is the relative position in the local ankle frame, $^{G}\mathbf{R}^{A}$ is the transform matrix from the local ankle frame $A$ to the inertial frame $G$, namely,
\begin{equation*}
  ^{G}\mathbf{R}^{A} =
\begin{bmatrix}
  \c_{\theta_A}& -\s_{\theta_A} \\
  \s_{\theta_A}& \c_{\theta_A}
\end{bmatrix}, \;
^{G}\dot{\mathbf{R}}^{A} =~^{G}\omega_A
\begin{bmatrix}
  -\s_{\theta_A}& -\c_{\theta_A} \\
  \c_{\theta_A}& -\s_{\theta_A}
\end{bmatrix},
\end{equation*}
where $^{G}\omega_A$ is the local ankle frame rotation angular velocity, $\theta_A$ is the ankle joint angle, and notations $\c_{\theta_A}=\cos \theta_A$ and $\s_{\theta_A}=\sin \theta_A$ for $\theta_A$. To obtain the position and corresponding velocity of point $p_i$, we can calculate the inverse kinematics of the robot leg in the local body frame $B$ such that $\theta_A=\pi/2-(\theta_1+\theta_2)$ and $^{G}\omega_A=~^{B}\omega_A=\omega_1+\omega_2$ for 2D problem. Furthermore, in this paper, we treat the robot walking gait as a given profile and the relative ankle position and velocity in the body-fixed frame, $B$, can be calculated in advance.

\subsubsection*{Kinetics}The ground reaction forces (GRFs) applied on foot feet are directly transmitted to the bipedal walker body with a mass, $M$. Denoting $\bs{r}_C=[x_{COM} \; z_{COM}]^T$ is the COM position, the kinetics of the COM is written as
\begin{equation}
\label{eqn:kinetics}
\ddot{\bs{r}}_C=\dot{\bs{v}}_C=\frac{1}{M}\mathbi{F},
\end{equation}
where $\bs{a}_C=\dot{\mathbi{v}}_C = [a_x \; a_z]^T$ is the acceleration of the COM in frame $G$ and $\mathbi{F} = [F_x \; F_z - Mg]^T$. $F_{x}$ and $F_z$ are the GRFs given by~\eqref{eqn:RFT_totalforce}. Given a small time step $\Delta t$, the velocity and position states of COM are updated by the explicit forward integration method,
\begin{equation}
\label{eqn:stateUpdate}
\bs{v}_C(t+\Delta t)=\bs{v}_C(t)+\bs{a}_C(t)\Delta t,
      \bs{r}_C(t+\Delta t)=\bs{r}_C(t)+\bs{v}_C(t)\Delta t.
\end{equation}


%
%
%
%
%
%

\section{Walking Energy Compensation}
\label{sec:walkingAnalysis}

\subsection{Instantaneous Power and Cumulative Work}

During the foot intrusion partially through granular terrain, the distributed resistive forces $f_{x}^i$ and $f_{z}^i$ on the foot perform negative work because the forces result in the opposite motion direction of the intruder. The bipedal walker needs to compensate for the negative work performed by reaction forces to maintain the normal gait. Therefore, the energy expense by the walker is a significant factor for the walking efficiency with given terminal states (e.g., the forward walking distance and average walking velocity). The power and work performed by resistive forces are calculated
\begin{equation}
\label{eqn:PowerandWork}
P_i =  \mathbi{f}_i \cdot \mathbi{v}_i, \; W = \int_0^t{\left(\sum_{i=1}^{N}P_i\right)dt}.
\end{equation}
where $\mathbi{f}_i=[f_x^i \;f_z^i]^T$, $\mathbi{v}_i=[v_{ix} \;v_{iz}]^T$ and $P_i$ are the force and velocity vectors and the instantaneous power of the $i$th plate, and $W$ is the cumulative work, respectively.
\setcounter{figure}{1}
\begin{figure}[t!]
	\hspace{-2mm}
	\subfigure[]{
	\label{fig:WalkingGait:a}
		\includegraphics[width=1.1in]{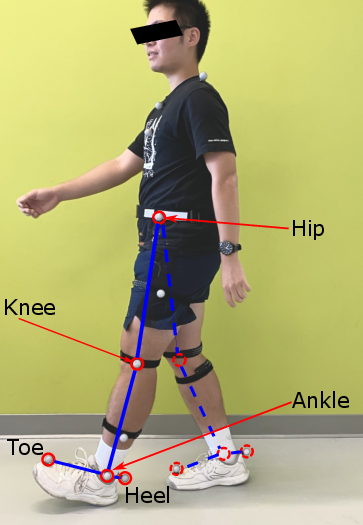}}
	\hspace{-2mm}
	\subfigure[]{
	\label{fig:WalkingGait:b}
		\includegraphics[width=2.28in]{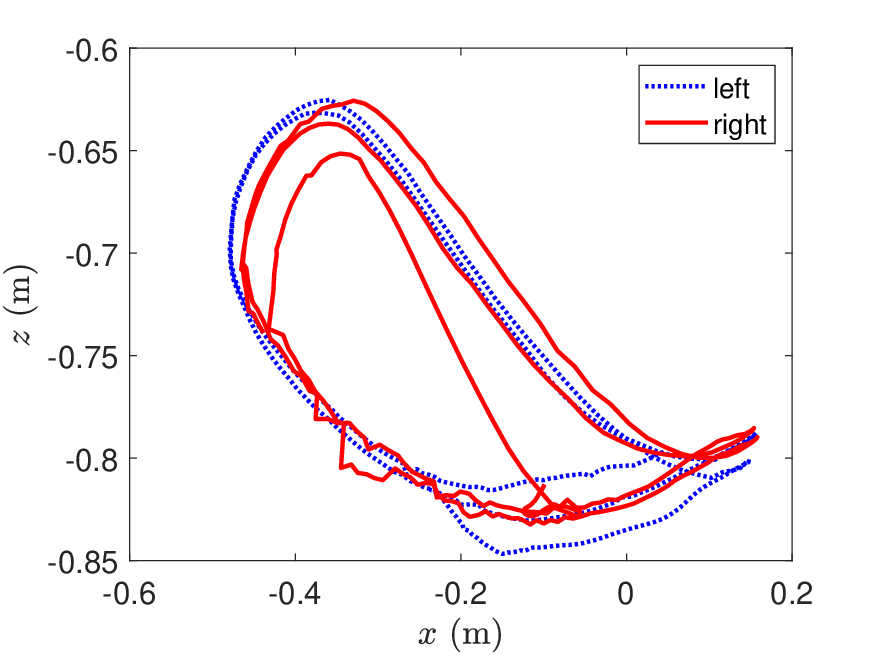}}
	\caption{(a) Human walking with optical markers on hip, knees, and ankles. (b) The motion profiles of the left and right ankle positions relative to the hip.}
	\label{fig:WalkingGait}
\end{figure}
\begin{figure}[h!]
	\centering
	\includegraphics[width=2.5in]{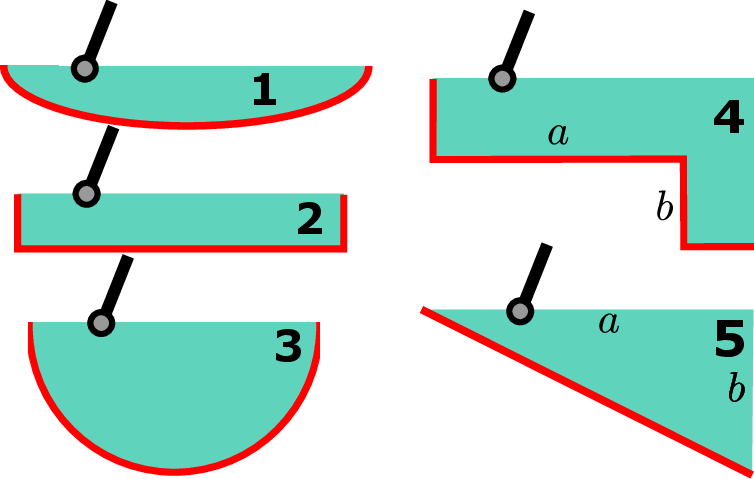}
	\caption{Five types of different foot shapes. 1: elliptical curve; 2: rectangle; 3: circular curve; 4: reversed-L; 5: reversed triangle.}
	\label{fig:pedalshapes}
\end{figure}
\setcounter{figure}{3}
\begin{figure*}[t!]
	\centering
	\subfigure[]{
		\label{fig:BodyStates_DifferentShapes}
		\includegraphics[width=2.9in]{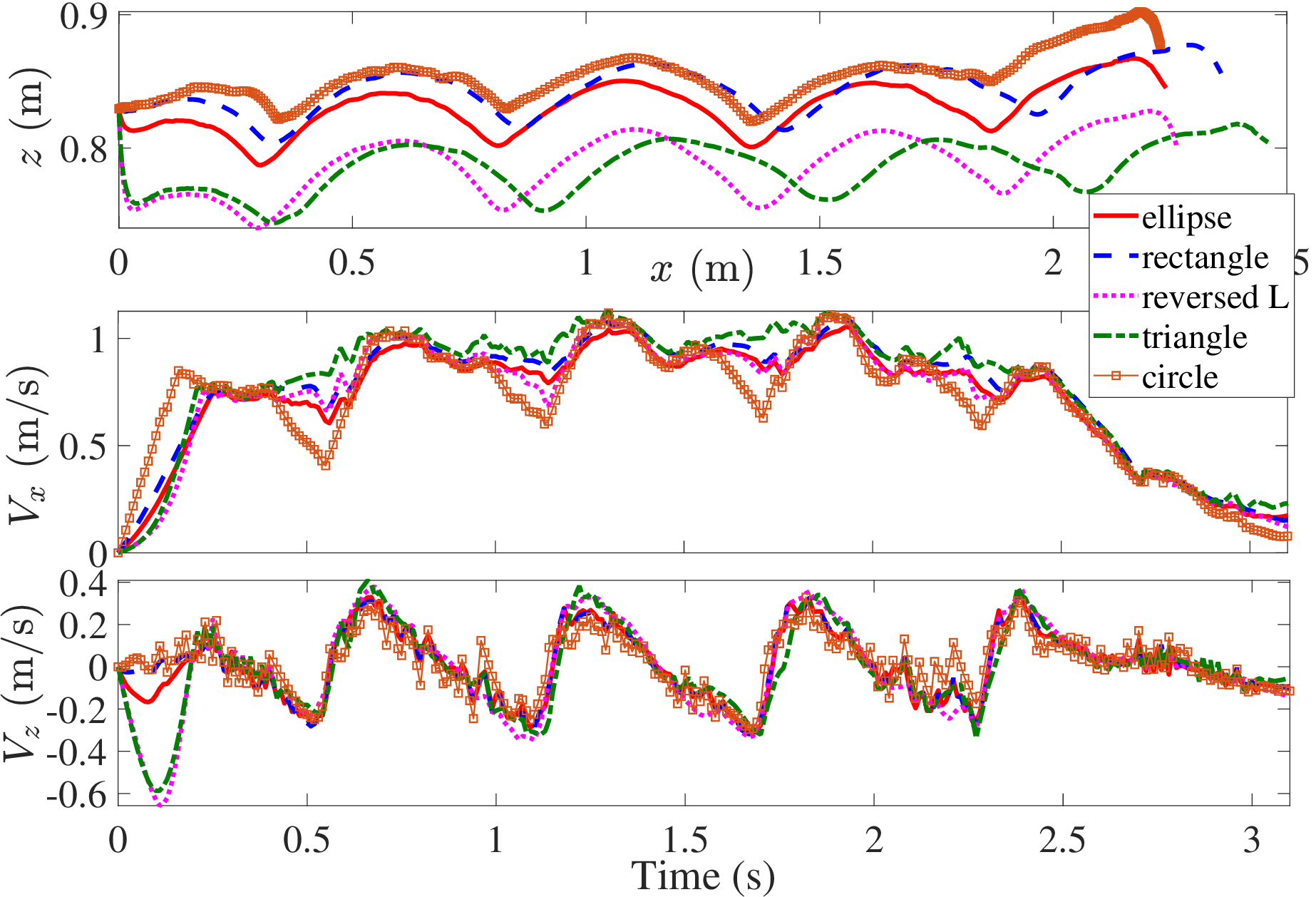}}
	\subfigure[]{
		\label{fig:PowerandWork_DifferentShapes}
		\includegraphics[width=2.9in]{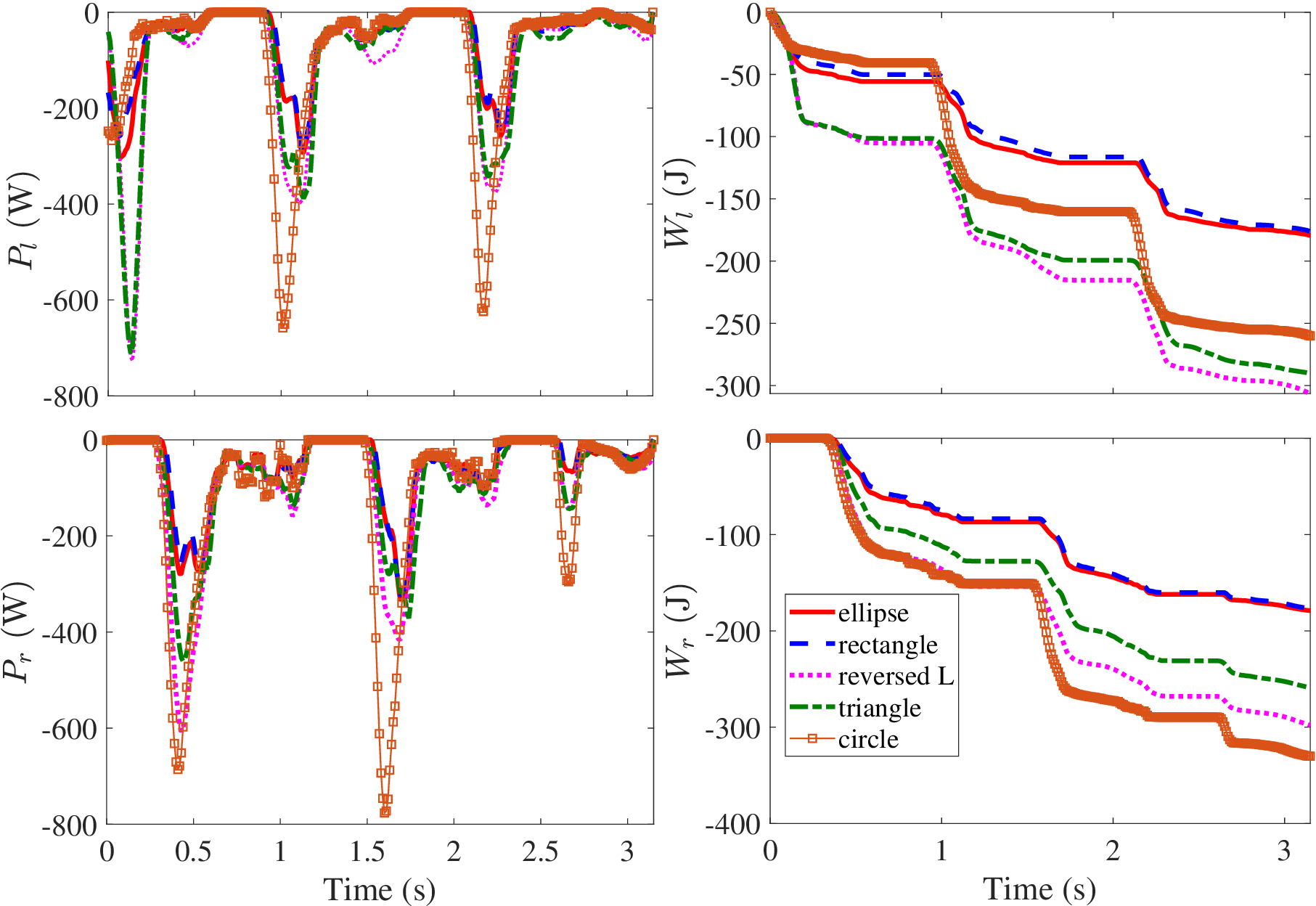}}
\vspace{-2mm}	
\caption{Comparison results under five different foot shapes. (a) The body COM position $\bs{r}_C$ and velocity $\bs{v}_C$ profiles. The first row is the $(x_{COM},z_{COM})$ and the second and third rows are horizontal $V_x$ and vertical velocity $V_z$, respectively. (b) Power ($P_l$ and $P_r$) and cumulative work ($W_l$ and $W_r$) by resistive forces on the left and right feet.}
\end{figure*}

\subsection{Walking Analysis of Different Foot Shapes}
Foot-granular terrain  intrusion depends on walking gaits. In this work, we use human walking gaits as the reference input to the intrusion forces calculation and then focus on the foot-shape impact on intrusion forces and resultant walker dynamics. Fig.~\ref{fig:WalkingGait:a} shows the experimental setup for human subject walking on the flat ground with motion capture systems to obtain the hip, knee, and ankle joint angles. The temporal ankle position profile in the global frame is then transferred to a spatial profile relative to the hip shown in Fig.~\ref{fig:WalkingGait:b}.

Five different foot shapes are selected in this study and Fig.~\ref{fig:pedalshapes} illustrates these shapes, including elliptical (e.g., C-shape~\cite{xu2015hybrid}), rectangle (e.g., flat foot~\cite{xiong2017stability}), circular (e.g.,~\cite{martin2011experimental}), reversed-L shape (\cite{wang2021continuum}), and triangle shape. These shapes are selected since they are previously reported in literature. To make a fair comparison, the effective foot-terrain contact lengths of all shapes, that is, the red lines in Fig.~\ref{fig:pedalshapes}, are the same. Similarly, to minimize the possible geometry effect, the aspect ratio (long and short axes) of the ellipse is set as $2:1$ and for other shapes, the ratio $a:b$ of side lengths $a$ and $b$ is also set as $2:1$.

We conducted simulation studies for a bipedal walker with mass $M=60$~kg and with the above five-mentioned foot shapes. Walker body position $\bs{r}_C$ and velocity $\bs{v}_C$ and the energy expenses to compensate for the intrusion forces are used as comparison metrics.  Fig.~\ref{fig:BodyStates_DifferentShapes} shows the COM position $\bs{r}_C$ and velocity $\bs{v}_C$ under five foot shapes. Fig.~\ref{fig:PowerandWork_DifferentShapes} shows the corresponding power and work for the left and right legs. From these results, we obtain the following observations.

First, among the five foot shapes, the walker with the triangular foot walks slightly fastest for the given ankle motion profile, followed by the flat foot as the next one. The walker with the circular foot however maintains the highest COM height and ends at the shortest forward walking distance. Second, the walker's horizontal velocity with the triangle feet is highest, while the circle feet give a significant velocity reduction at each stance switch. No obvious vertical velocity difference is observed among these five foot shapes. Finally, the walker with circular feet needs the highest power during the single stance phase while elliptical and rectangular feet have the least. Therefore, by given walker's gait profile, the gait stability and energy efficiency depend on the foot shapes and their intrusion on granular terrains.

\section{Optimal Foot Contour Design}
\label{sec:optimalDesign}

\subsection{Waypoints of Foot Contour}
Fig.~\ref{fig:waypoints} illustrates the schematic of the walker's foot contour under a set of waypoints, which allows flexibility to adapt the shape corresponding to intrusion interactions. Considering the real foot-terrain interaction, granular materials intrusion mainly happens at the foot bottom and a set of $n$ waypoints, denoted by $p_i$, $i=1,\ldots,n$, are set up at the bottom contour. The foot contour is restricted in a rectangular domain ($2L\times2H$, gray shadowed area) with a local foot frame $x_po_pz_p$ at the center.
For simplicity, waypoints $p_i$, $i=1,\ldots,n$, are located uniformly in the $x_p$ direction while the $z_p$ coordinates of points are free bounded in the domain. A cubic spline is then used to connect any two adjacent points and the coordinates of $p_i$ are
\begin{equation}
\label{eqn:waypoints}
\bs{p}_i = \left(\frac{(i-1)(3-n)}{n-1}L,~-k_i\frac{H}{K}\right),\;i=1,\ldots,n,
\end{equation}
where $K \in \mathbb{N}$ is the total number of discretization of the height $H$ and $k_i\in \mathbb{Z_+},k_i\leq K$ is the index location of point $p_i$  in the $z_p$ direction. Therefore, for each $p_i$, we just need to determine value of $k_i$s and this can be formulated as an integer optimization problem. We define design variable $\bs{k}=[k_1 \; k_2 \; \cdots \,k_n]^T$ for the optimization problem that will be introduced later in this section.

\setcounter{figure}{4}
\begin{figure}[h!]
  \centering
  \includegraphics[width=2.6in]{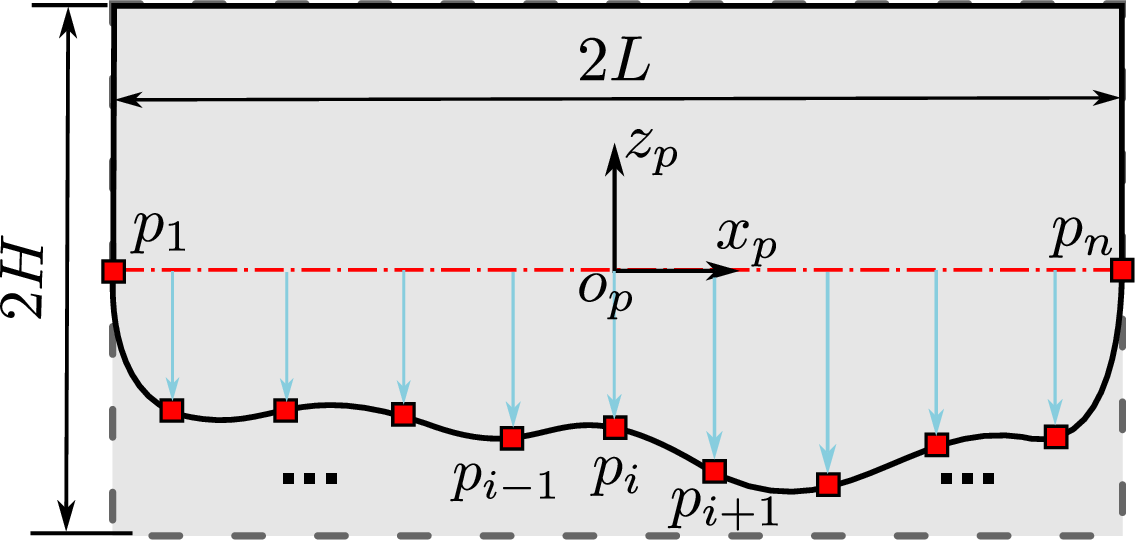}
  \caption{The schematic of the walk foot with finite waypoints to control the contour. The foot shape is controlled by a set of waypoint location $\bs{p}_i$.}
  \label{fig:waypoints}
\end{figure}

\subsection{Optimization of Foot Contour}

We formulate an optimization problem with the following considerations. First, for given a certain period $t_f$, the bipedal walker should walk for a as long distance as possible, that is, $x_{COM}(t_f)$ is large as possible. Second, the foot sinkage should be as small as possible because drag force and compensation work arise as the intrusion depth increases. We evaluate the sinkage depth through the (average) height of $z_{COM}(t)$. Third, during a single-stance walking phase, the instantaneous power peak $p_{\max}=\max{\left(\sum_{i=1}^{N}P_i\right)}$ should be small for reducing impact effect. The work $|W(t_f)|$ by the drag and lift forces should be small to save energy. Finally, the walker should not lose too much forward velocity change $\Delta V_x$ when the walker switches the stance legs.

\setcounter{figure}{6}
\begin{figure*}[t!]
  \centering
  \includegraphics[width=6.9in]{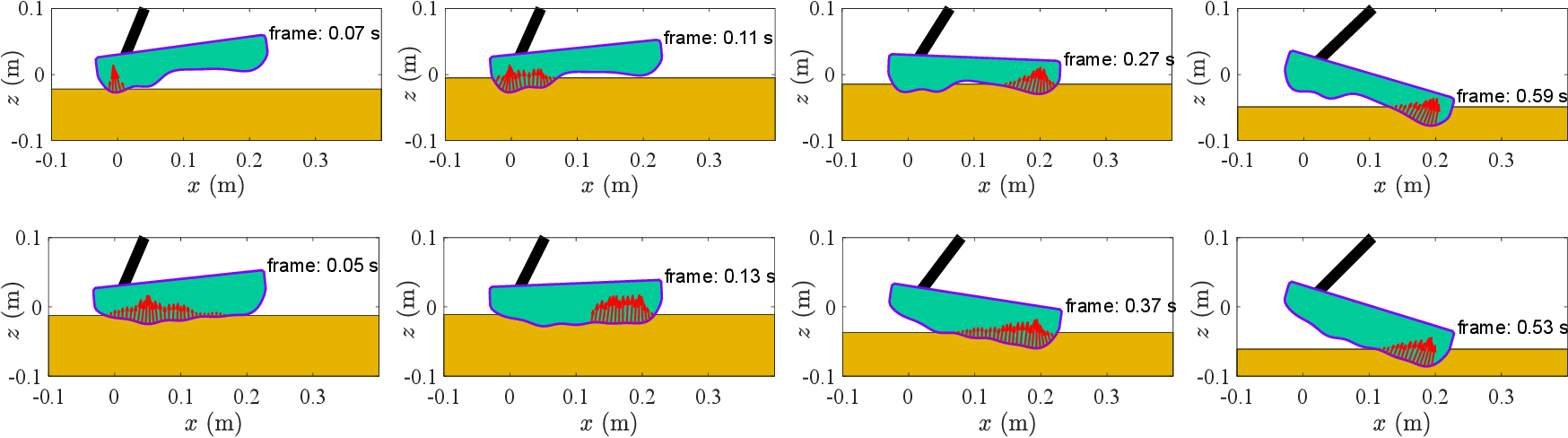}
  \vspace{-1mm}
  \caption{One step on hard sand from touching down to pushing off. The first row is the optimal foot shape of hard sand (red dashed line in Fig.~\ref{fig:optimalShapes}) and the second row is the optimal shape of regular sand (green solid line in Fig.~\ref{fig:optimalShapes}). Red arrows reflect the intrusion force distribution showing the directions and magnitudes.}\label{fig:footevolution}
\end{figure*}

The cost function, denoted by $J_W$, that reflects aforementioned objectives is chosen as a form of positive and negative rewards as
\begin{equation}
\label{eqn:costfcn}
J_W = -\frac{R_p}{R_n},
\end{equation}
where $R_p$ and $R_n$ indicates the positive and negative rewards, respectively,
\begin{equation}
\label{eqn:reward}
R_p= x_{COM}(t_f) \bar{z}_{COM}(t),\; R_n= |W(t_f)| p_{\max} \Delta V_x,
\end{equation}
and $\bar{z}_{COM}(t)$ is the mean value of ${z}_{COM}(t)$ over period $t_f$. Therefore, the optimization problem for the foot-shape design is formulated as
\begin{equation}\label{eqn:optimization}
  \begin{aligned}
    &\min_{\bs{k}} ~ J_W \\
    \text{subj} \; & \text{to:} ~~k_i\in \mathbb{Z_+},~k_i\leq K, \, i=1,\ldots,n.
  \end{aligned}
\end{equation}
The cost function $J_W$ can be viewed as an index of walking efficiency since the term $\frac{x_{COM}(t_f)}{|W(t_f)|}$ actually reflects the walking distance per unit work. The larger is this value, the more efficiently the robot walks on the granular terrain. The optimization problem~\eqref{eqn:optimization} is solved by genetic algorithm for the mixed integer optimization. We will present the optimal results in the next section.
\section{Results and Discussions}
\label{sec:Results}
\subsection{Simulation Setup}
We first determine the simulation model parameters such as foot shape domain size, the domain grid, and the number of intrusion plates. We consider a human walker in the simulation study. The bipedal walker first stands with the left leg steadily and steps forwards following the given human walking gait profile on granular terrains. The simulation ends at a terminal time $t_f$. Table~\ref{tab: simParas} lists the values of the simulation parameters used in this study.

\renewcommand{\arraystretch}{1.3}
\begin{table}[b!]
\centering
\caption{The parameter values for the bipedal walkers and foot shape.}
\label{tab: simParas}
\resizebox{\columnwidth}{!}{%
\begin{tabular}{cccccccc}
\hline
$L$    & $H$    & $n$ & $K$ & $N$ & $M$   & $l_1$  & $l_2$  \\ \hline
$0.13$~m & $0.03$~m & $11$  & $10$  & $100$ & $60$~kg & $0.47$~m & $0.45$~m \\ \hline
\end{tabular}%
}
\end{table}



\setcounter{figure}{5}
\begin{figure}[t!]
  \centering
  \includegraphics[width=2.5in]{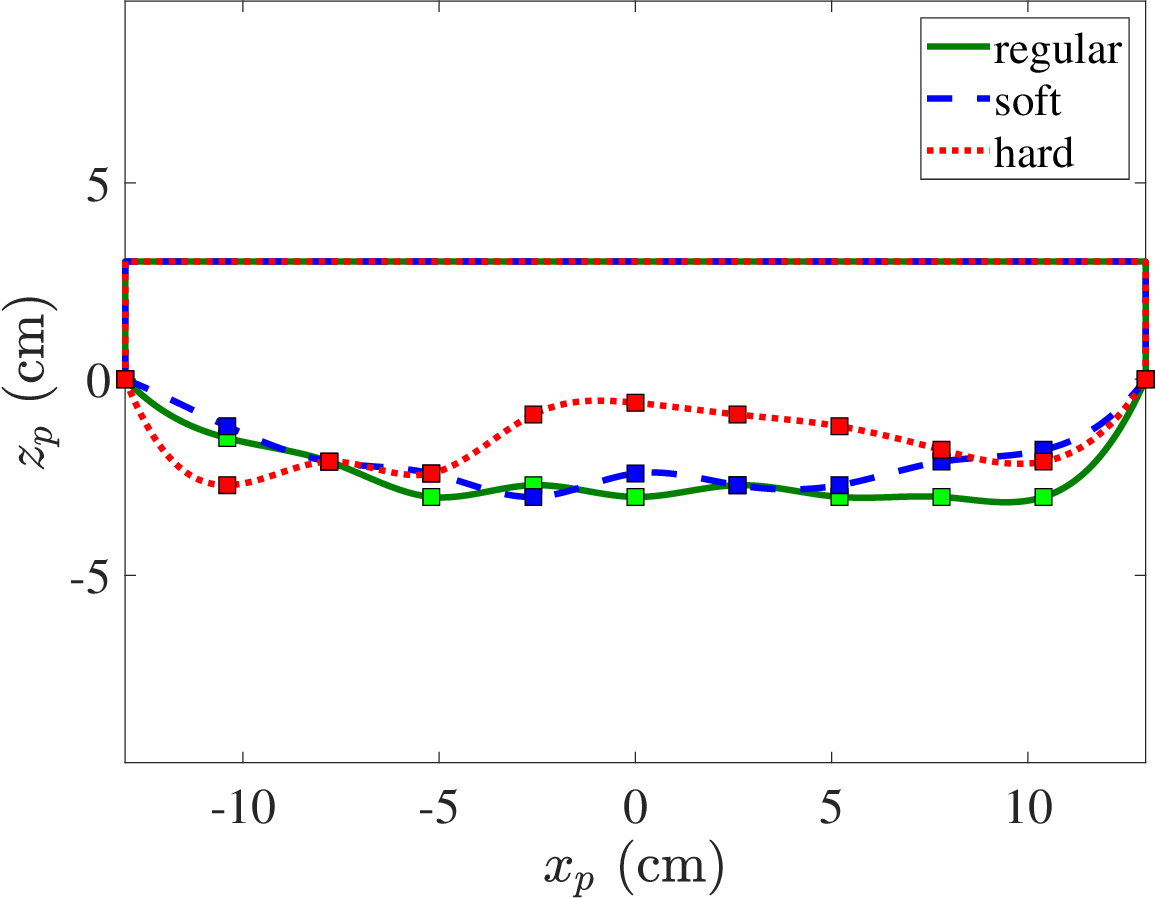}
  \vspace{-2mm}
  \caption{Three optimal foot contours for walking on three different types of sand.}
  \label{fig:optimalShapes}
\end{figure}

\setcounter{figure}{7}
\begin{figure*}[t!]
	\hspace{-3mm}
	\subfigure[]{
		\label{fig:results_BodyPos_Vel}
		\includegraphics[width=2.32in]{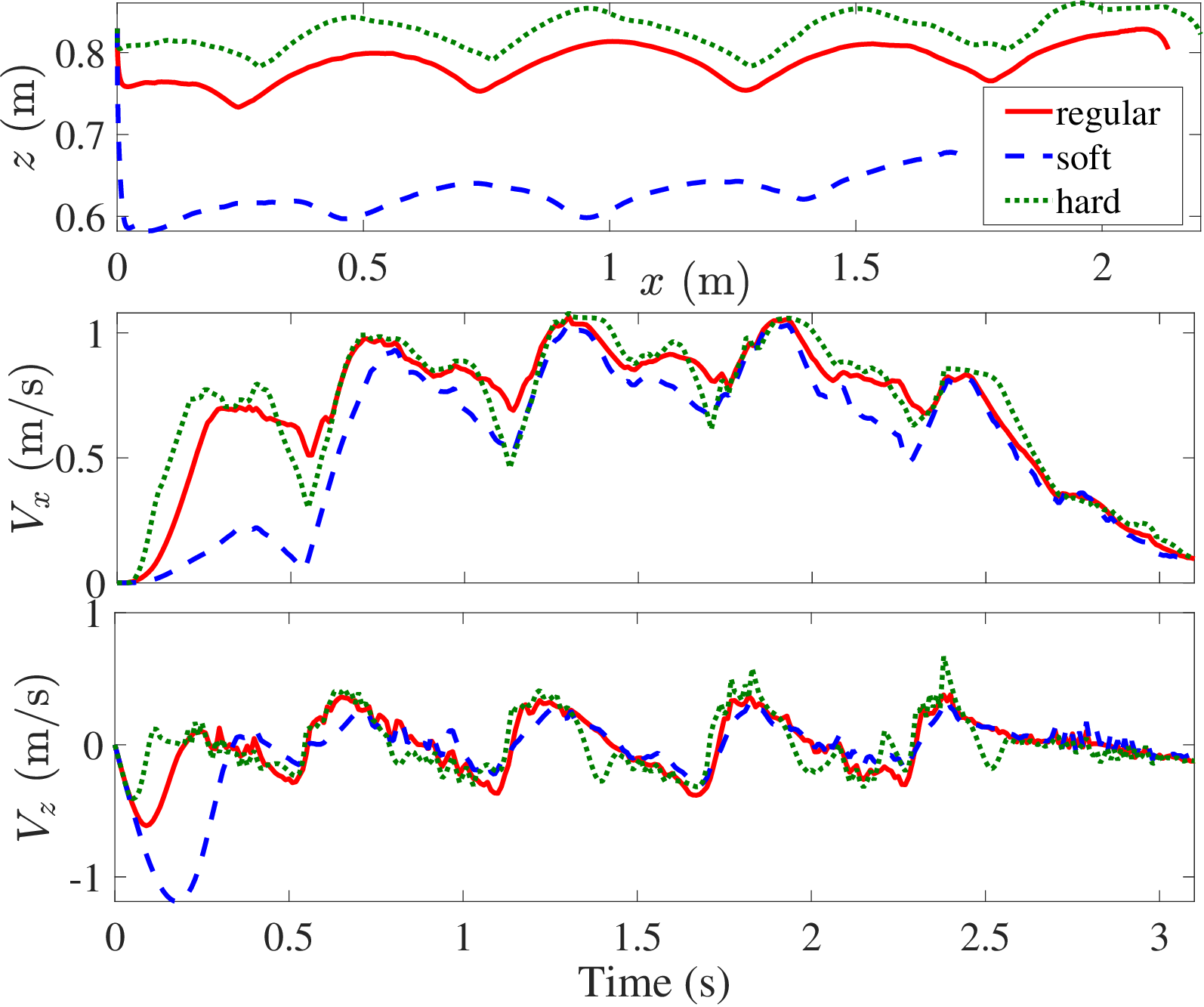}}
	\hspace{-1mm}
	\subfigure[]{
		\label{fig:results_Forces}
		\includegraphics[width=2.35in]{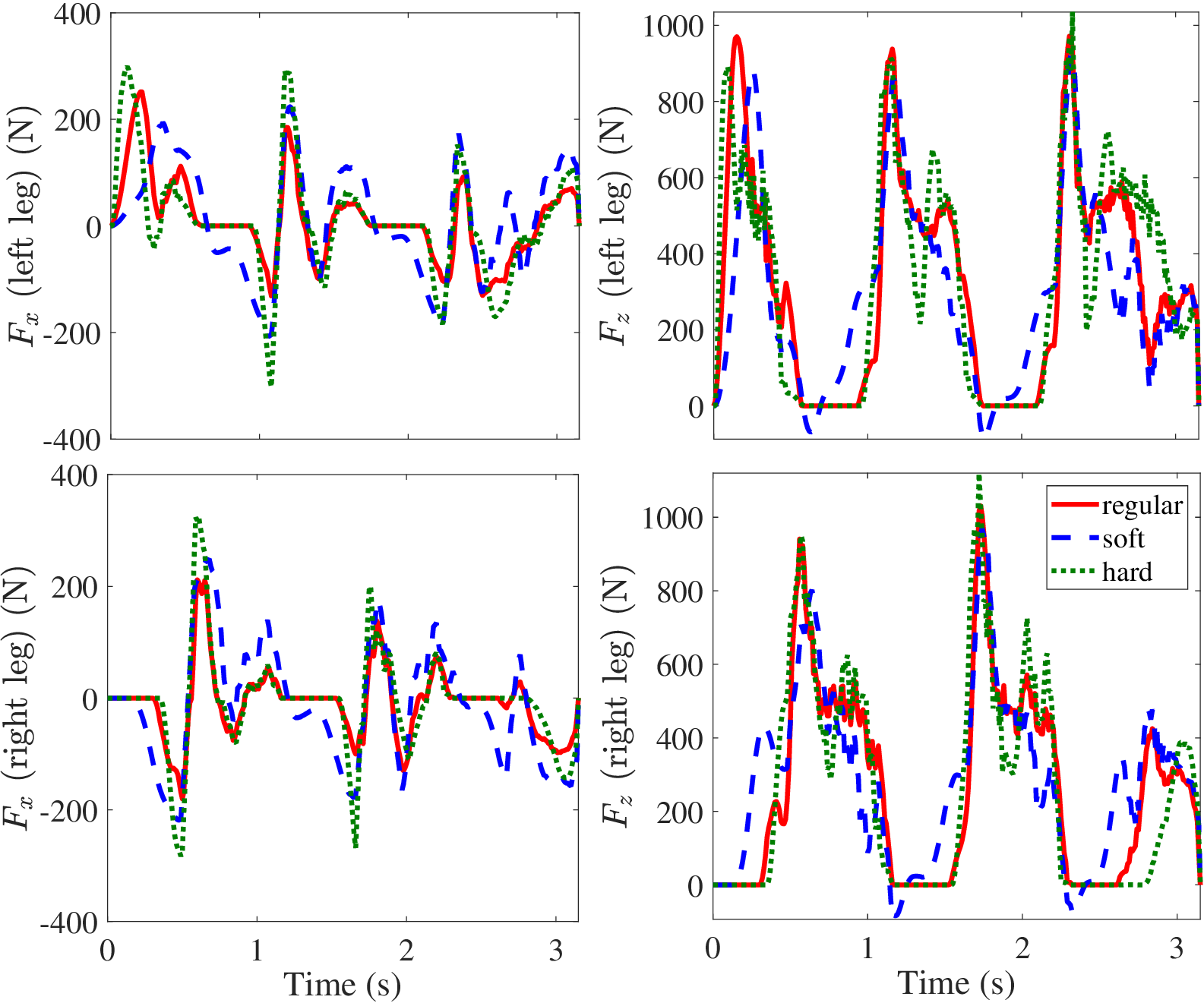}}
	\hspace{-1mm}
	\subfigure[]{
		\label{fig:results_PowerWork}
		\includegraphics[width=2.38in]{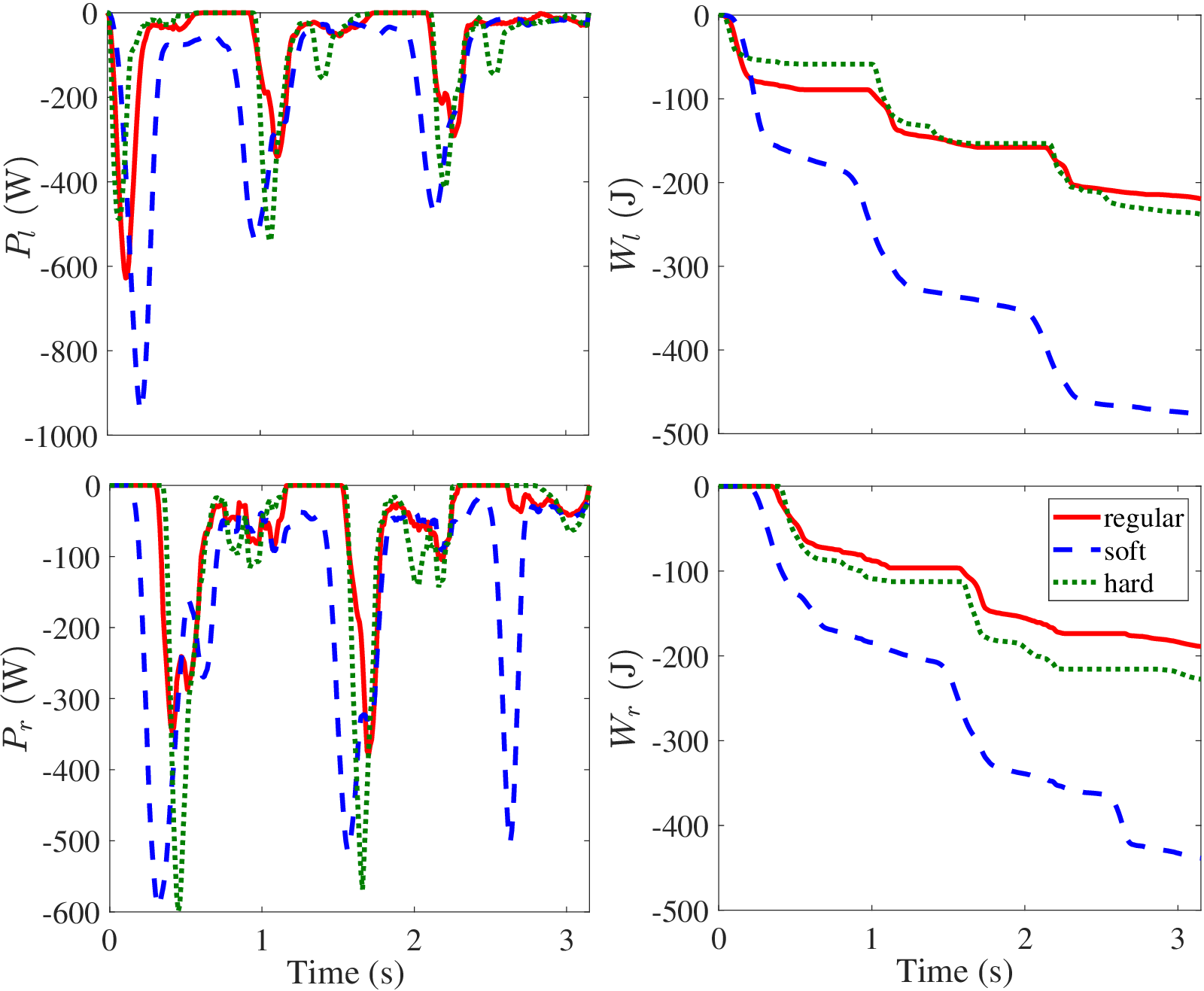}}
\vspace{-1mm}	
\caption{(a) The COM position and velocities on three different types of sand. (b) The ground reaction forces during walking on the sand. The left column: drag force; the right column: lift (normal) force. (c) Instantaneous power and cumulative work compensation caused by sand resistive forces on right legs. The left column: instantaneous power; the right column: work.}
\end{figure*}

Three different types of sand are tested for analysis of the impact on the foot contour and walking dynamics. We label ``regular'', ``soft'', and ``hard'' to describe the intrusion interaction type of the foot with sand. The intrusion interaction is characterized by the local stress maps that are described by $\alpha_{x}(\beta,\gamma)$ and $\alpha_{z}(\beta,\gamma)$. One simple scale factor $\zeta$ is used to distinguish different types of granular materials. The specific local stress $\alpha_{j}= \zeta \alpha_{j}^{\mathrm{generic}}$, $j=x,z$, where $\alpha_{j}^{\mathrm{generic}}$ is the local stress of generic materials discussed in~\cite{zhang2014effectiveness}. In this paper, $\zeta_{r} = 1$, $\zeta_{s}=0.2$, and $\zeta_{h}=5$ are for regular, soft, and hard sand, respectively. We chose these three numbers primarily to significantly distinguish the local stress difference in the results.

\subsection{Optimal Foot Contours on Different Types of Sand} Fig.~\ref{fig:optimalShapes} shows the optimal foot contour results for different types of sand.
It is noted that for walking on the hard sand the optimal foot shape appears unnecessarily convex, while the contour tends to be the curve with a small curvature for walking on regular and soft sand. This is partially because the small-curvature shape provides large effective contact zone to avoid significant sinkage. Fig.~\ref{fig:footevolution} illustrates the process of the foot stepping on the hard sand with foot shapes optimized for hard and soft sands. From (heel) touch-down to (toe) push-off, the concave shape (red dashed line in Fig.~\ref{fig:optimalShapes}) reduces unnecessary contact zone in the middle area of the foot. This leads to less numbers of intrusion plates and saves the work. From the force distribution results, we can see that the step pattern of the single stance with the concave foot shape is similar to the human walker: the force moving the body forwards is mostly provided in the (toe) push-off phase (e.g., $t>0.27$ s). However, for the ``overall convex'' curved foot shape like the green solid line in Fig.~\ref{fig:optimalShapes}, a larger contact zone is involved during the interaction and shifts to the leading part of the foot, which reveals that the duration of the push-off phase becomes longer (good for increasing the forward velocity) while larger instantaneous power and work are generated to be compensated by the joint actuators.

Fig.~\ref{fig:results_BodyPos_Vel} shows the walker's COM position and velocity profiles. The walker can walk furthest over the same time period on the hard sand that is closed to normal firm ground, while the sinkage level is the smallest among three types of sand terrain. The soft sand leads to the largest sinkage and shortest walking distance. From velocity profiles, the walker experiences an obvious loss of the horizontal velocity at the time when changing the stance leg (i.e., at $t=1.1$, $1.7$, and $2.3$~{s} on the soft sand. Meanwhile, the walker's acceleration on the soft sand is also smallest, resulting more time to reach the normal walking velocity around $1$~m/s. Fig.~\ref{fig:results_Forces} shows the GRF profiles (intrusion drag and lift forces). The larger sinkage on the soft sand results in larger drag forces in both the fore and aft directions (blue dashed line) and longer contact/intrusion phase compared with the results on regular and hard sand. Fig.~\ref{fig:results_PowerWork} shows the corresponding instantaneous power and cumulative work of both legs. After three steps, the cumulative work on soft sand is largest. All results demonstrate that walking on soft granular materials needs energy to compensate for resistive forces as well as a short walking distance with a slow velocity, which implies that soft terrain reduces the walking efficiency.



\section{Conclusions and Future Work}
\label{sec:conclusion}

We presented a computational approach to calculate the ground reaction forces during the interaction of bipedal walker feet with dry granular terrains. The method first determined the foot/terrain intrusion interaction motion. We then quantified five different foot shapes by the walker's COM motion profile, the ground resistive forces and the energy efficacy. A multi-objective optimization problem for foot shape design was formulated to minimize the work compensation for intrusion forces. Different types of sand were used for testing the walking gait profiles with the optimized foot shapes. The simulation results showed that the non-convex curved foot shape generated the best energy efficiency for the human walking gait on hard granular terrain. One ongoing research task is to conduct experimental validation of the proposed interaction model. Furthermore, human walking locomotion with assistance of knee exoskeletons~\citep{Chen2021TMECH} and gait trajectory (including velocity) effect will be investigated for advanced bipedal locomotion control on granular terrains.

\bibliography{MECC2023_Ref}

\end{document}